\documentclass{article}
\UseRawInputEncoding
\usepackage{microtype}
\usepackage{graphicx}
\usepackage{subfigure}
\usepackage{booktabs} 

\usepackage{caption} 
\usepackage{needspace}   
\usepackage{stfloats}   
\usepackage[section]{placeins} 
\usepackage{amsmath}
\usepackage{amssymb}
\usepackage{mathtools}
\usepackage{amsthm}
\usepackage[capitalize,noabbrev]{cleveref}
\usepackage[textsize=tiny]{todonotes}
\usepackage[accepted]{icml2024}

\icmltitlerunning{Scaling Graph Transformers: Sparse vs. Dense Attention}

\begin{document}

\twocolumn[
\icmltitle{Scaling Graph Transformers: A Comparative Study of Sparse and Dense Attention}

\icmlsetsymbol{equal}{*}

\begin{icmlauthorlist}
\icmlauthor{Leon Dimitrov}{tum}
\icmlaffiliation{tum}{Department of Computer Science, Technical University of Munich, Munich, Germany}
\icmlcorrespondingauthor{Leon Dimitrov}{leon.dimitrov@tum.de}
\end{icmlauthorlist}

\icmlkeywords{Graph Transformers, Attention, Sparse Attention, Dense Attention, Scalability}

\vskip 0.3in
]

\renewcommand{\thefootnote}{}%
\footnotetext{
\textsuperscript{1}Independent, Munich, Germany. Correspondence to: Leon Dimitrov \textless leon.dimitrov@tum.de\textgreater.
}
\setcounter{footnote}{0}
\renewcommand{\thefootnote}{\arabic{footnote}}

\begin{abstract}
Graphs have become a central representation in machine learning for capturing relational and structured data across various domains. Traditional graph neural networks often struggle to capture long-range dependencies between nodes due to their local structure. Graph transformers overcome this by using attention mechanisms that allow nodes to exchange information globally. However, there are two types of attention in graph transformers: dense and sparse. In this paper, we compare these two attention mechanisms, analyze their trade-offs, and highlight when to use each. We also outline current challenges and problems in designing attention for graph transformers.
\end{abstract}

\section{Introduction}
\label{sec:intro}

Graphs naturally represent structured data as nodes and edges, making them a core tool in machine learning applications ranging from molecular biology to social networks~\cite{kipf2017semi}. Graph neural networks (GNNs) are widely used for this purpose, relying on local message passing to share information between neighboring nodes~\cite{gilmer2017neural}. However, their locality often limits their ability to model long-range dependencies, which are crucial in many real-world tasks~\cite{dwivedi2020benchmarking,alon2020oversquashing}.

To address these limitations, graph transformers extend the self-attention \mbox{mechanism-originally} developed for natural language processing~\cite{vaswani2017attention}-to graph-structured data~\cite{ying2021graphormer,kreuzer2021rethinking}. Unlike traditional GNNs that rely on local message passing, transformers allow each node to attend to all others, enabling flexible aggregation of global context. This is achieved through self-attention, which computes learned, weighted interactions between all node pairs. However, this global interaction introduces a trade-off between two attention strategies:
\textbf{Dense attention} offers full context at quadratic computational cost, while
\textbf{sparse attention} reduces complexity to linear by limiting connections, potentially at the cost of missing important long-range dependencies~\cite{rampasek2022graphgps}.

This trade-off between scalability and expressivity motivates our central question: \emph{When and how should graph transformers use dense versus sparse attention?}

We compare these paradigms by analyzing attention mechanisms, reviewing state-of-the-art models, summarizing empirical performance, and outlining theoretical gaps and ideal use cases.

\section{Background \& Conceptual Overview}
\label{sec:background}

To ground our comparison of dense and sparse attention mechanisms in graph transformers, we first review the standard self-attention operation and how it is adapted to graphs.

\subsection{Transformer Self-Attention Recap}
The Transformer architecture was originally introduced for sequence modeling tasks in natural language processing (NLP), such as machine translation \cite{vaswani2017attention}. Its core innovation is the self-attention mechanism, which enables each token in a sequence to dynamically weigh and incorporate information from all other tokens.

Formally, given an input sequence of $N$ token embeddings $X \in \mathbb{R}^{N \times d}$, the model computes queries, keys, and values as $Q = XW_Q$, $K = XW_K$, and $V = XW_V$, where $W_Q$, $W_K$, and $W_V$ are learned projection matrices. The attention weights are derived from the similarity between queries and keys, and the result is a weighted sum over the values.

The scaled dot-product attention is computed as:
\begin{equation}
\mathrm{Attention}(Q,K,V) = \mathrm{softmax}\left(\tfrac{QK^\top}{\sqrt{d_k}}\right)\,V
\end{equation}
This produces an updated representation for each token as a weighted combination of all token values in the sequence.

In standard self-attention, the attention matrix has size \( N \times N \), enabling full pairwise interactions and allowing the model to capture complex, long-range dependencies.

\subsection{Graph Transformer Basics}
Unlike sequences or grids, graph data lacks a natural ordering of elements. This poses a challenge for transformers, which rely on positional relationships between tokens. For graph-structured inputs, the model must be permutation-invariant, meaning its output remains unchanged when node indices are reordered \cite{dwivedi2020benchmarking}.

To address this, graph transformers inject structural bias using positional or structural encodings, such as Laplacian eigenvector embeddings, shortest-path distances, node degrees, or edge-type features~\cite{rampasek2022graphgps, kreuzer2021rethinking}.
Such encodings are typically added to the input embeddings or incorporated into the attention mechanism \cite{ying2021graphormer, rampasek2022graphgps}, allowing the model to reason about node identity and relative position within the graph topology.

Unlike message-passing GNNs, which propagate information locally along edges, graph transformers enable  global communication between all nodes in a single layer. This facilitates modeling of long-range dependencies but incurs computational cost on large graphs, as attention weights must be computed for every node pair~\cite{shirzad2023exphormer}.

\subsection{Dense vs. Sparse Attention}
The design of a graph transformer fundamentally hinges on how attention is distributed across nodes. While self-attention provides a powerful framework for capturing global dependencies, it can be implemented with varying levels of connectivity.

\textbf{Dense attention} mechanisms maintain full pairwise interactions across all nodes, preserving maximum expressive capacity and enabling every node to consider the entire graph context at each layer. However, this comes at the cost of quadratic scaling with respect to the number of nodes, both in time and memory~\cite{ying2021graphormer}.

\textbf{Sparse attention}, by contrast, restricts the attention computation to a subset of node pairs. These subsets may be based on the original graph structure (e.g., adjacent nodes), predefined sparsity patterns, or dynamically learned schemes~\cite{shirzad2023exphormer, rampasek2022graphgps}. Sparse designs can reduce the complexity to linear, often without a substantial loss in performance, especially on large graphs.

\begin{figure}[H]
    \centering
    \includegraphics[width=0.85\linewidth]{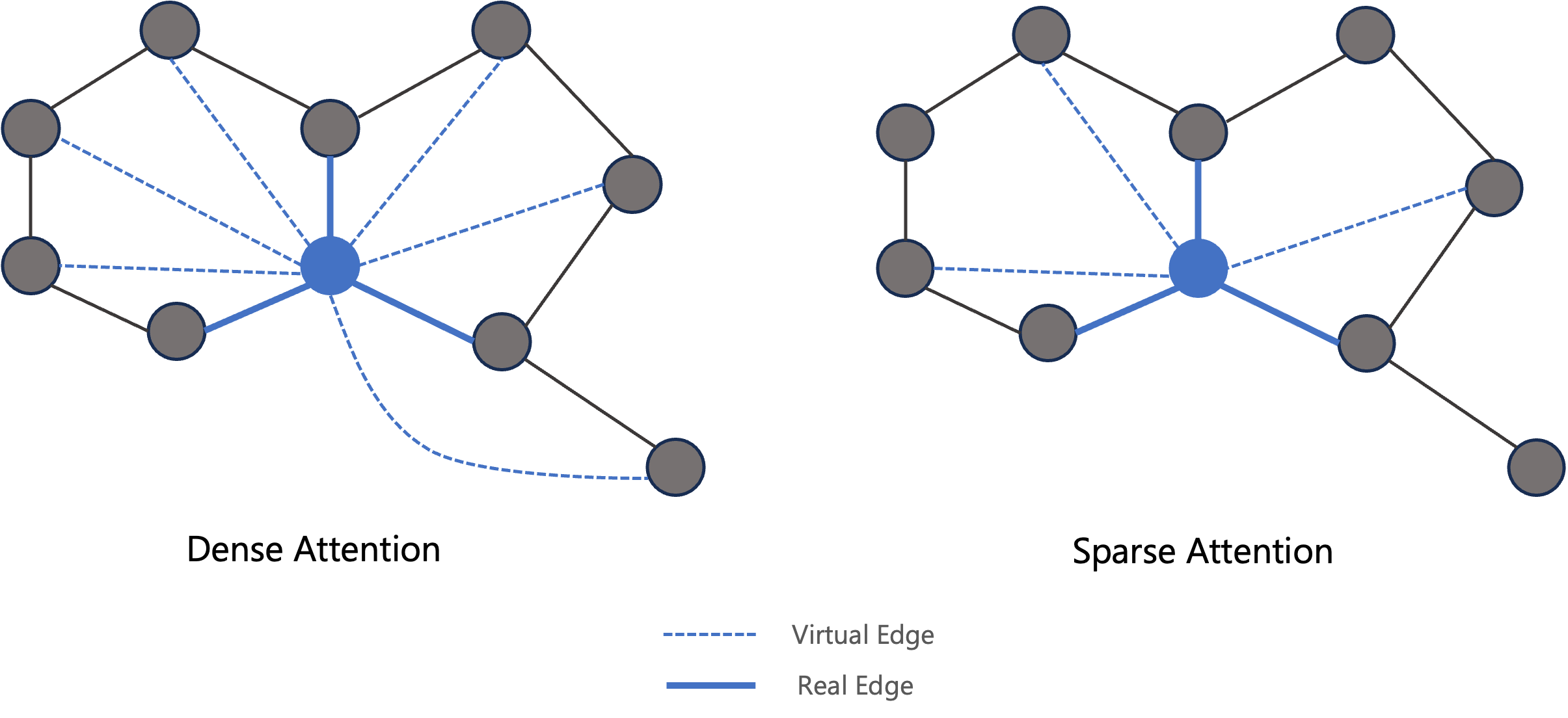}
    \caption{Comparison of dense vs. sparse attention}
    \label{fig:dense-vs-sparse}
\end{figure}

This design choice affects both scalability and how information flows through the graph. Dense models capture global context in a single layer, while sparse models rely on deeper stacks to achieve similar reach. Figure~\ref{fig:dense-vs-sparse} illustrates this distinction: in the dense setting, each node attends to all others, while sparse attention restricts each node's attention to a limited subset of other nodes.

\vspace{-1.0em} 
Next, we examine representative models that implement these paradigms and compare their design, efficiency, and performance.

\section{Key Methods \& Recent Research}

\subsection{Dense Graph Attention Models}

We begin with dense graph transformers, which implement full pairwise attention and often incorporate graph structure through learned biases in both the input embeddings and attention scores.

\textbf{Graphormer} \cite{ying2021graphormer} is a representative dense model that integrates graph topology directly into both its input representations and attention mechanism. The input embedding of each node \( i \) is augmented with a learnable encoding of its in-degree:
\vspace{-3pt}
\[
X_i \gets X_i + \mathrm{Embed}(\mathrm{deg}(i)),
\]
where \( \text{deg}(i) \) is the in-degree of node \( i \) and \( \text{Embed} \) is a degree embedding table. This introduces a simple but effective structural bias that allows the model to distinguish between locally important and peripheral nodes.

Graphormer also adds two learned biases to the attention logits: \(B_{\mathrm{sp}}[i,j]\), a scalar from a table indexed by the shortest-path distance \(d_G(i,j)\), and \(B_{\mathrm{edge}}[i,j]\), an embedding of the multiset of edge types along that path. The attention is then computed as~(2):
\begin{equation}
\mathrm{Attention}(Q,K,V) = \mathrm{softmax}\left(\frac{QK^\top}{\sqrt{d_k}} + B_{\mathrm{sp}} + B_{\mathrm{edge}}\right)V
\tag*{}
\stepcounter{equation}
\end{equation}
This architecture enables Graphormer to incorporate global structural information directly into attention scores, without using explicit positional encodings in the input.

Another notable dense-attention model is the \textbf{Spectral Attention Network (SAN)}~\cite{kreuzer2021rethinking}, which leverages Laplacian eigenpairs to build \emph{learned spectral positional encodings} that are \emph{concatenated} to the node embeddings before applying transformer layers.

Let \( G = (V, E) \) be an undirected graph with \( N \) nodes, adjacency matrix \( A \in \mathbb{R}^{N \times N} \), and degree matrix \( D \). The \emph{normalized graph Laplacian} is defined as \newline \( L = I - D^{-1/2} A D^{-1/2} \), where \( I \) is the identity matrix.

The matrix \( L \) is symmetric and admits an eigendecomposition \( L = \Phi \Lambda \Phi^\top \), where \( \Phi \) contains orthonormal eigenvectors and \( \Lambda \) is a diagonal matrix of non-negative eigenvalues.

In analogy to signal processing, Laplacian eigenvalues represent graph frequencies: small eigenvalues correspond to smooth, low-frequency patterns, while large ones reflect rapid variation. SAN uses the corresponding low-frequency eigenvectors to compute node similarity, favoring nodes that align in these globally coherent structures.

Concretely, SAN selects the first $K$ eigenpairs $\{(\lambda_k,\phi_k)\}_{k=1}^K$. For each node $i$, it forms a sequence
\begin{equation}
s_i = \big[(\lambda_1,\phi_1(i)),\, (\lambda_2,\phi_2(i)),\, \dots,\, (\lambda_K,\phi_K(i))\big],
\end{equation}
feeds $\{s_i\}$ through a lightweight \emph{Positional Encoding Transformer} to obtain a learned positional encoding $p_i \in \mathbb{R}^{d_p}$, and concatenates this with the original node feature $x_i$ to produce $\tilde{x}_i = [x_i \,\|\, p_i]$.

The graph transformer then computes attention over all node pairs using these augmented embeddings, with the standard scaled dot-product.
By learning positional encodings from low-frequency Laplacian eigenvectors and appending them to the node features, SAN injects global spectral structure while retaining the flexibility of conventional transformer attention.

\subsection{Sparse Attention Techniques}

To improve scalability, sparse attention limits each node’s interactions to a predefined or learned subset-such as neighbors or selected anchors-reducing complexity while preserving the ability to model long-range dependencies.
\begin{table*}[!b]
\setlength{\tabcolsep}{2pt}
\renewcommand{\arraystretch}{0.72}  
\centering
\tiny                                

\begin{tabular*}{\textwidth}{@{\extracolsep{\fill}}lccccl@{}}
\toprule
\textbf{Model} & \textbf{Type} & \textbf{Node Encoding} &
\textbf{Attention} & \textbf{Complexity} \\ \midrule

Graphormer & Dense &
$X_i + \mathrm{Embed}\!\bigl(\deg(i)\bigr)^{\dagger}$ &
$\operatorname{softmax}\Bigl(\tfrac{QK^\top}{\sqrt{d_k}}+B_{\mathrm{sp}}+B_{\mathrm{edge}}\Bigr)V$
& Quadratic \\

SAN & Dense &
$[X_i \,\|\, p_i]$ &
$\operatorname{softmax}\!\Bigl(\tfrac{QK^\top}{\sqrt{d_k}}\Bigr)V$ &
Quadratic \\

Exphormer & Sparse &
$X_i + P_i + V_{\mathrm{virt}}$ &
$\operatorname{softmax}\Bigl(\tfrac{QK^\top}{\sqrt{d_k}}\Bigr)_{\mathcal{N}_i}V_{\mathcal{N}_i}$
& Linear \\

GraphGPS & Sparse &
$[\,X_i \,\|\, \mathrm{PE}_i \,\|\, \mathrm{SE}_i\,]$ &
$\operatorname{softmax}\Bigl(\tfrac{QK^\top}{\sqrt{d_k}}\Bigr)_{\mathcal{N}_i}V_{\mathcal{N}_i}$
& Linear \\
\bottomrule
\end{tabular*}

\caption{Comparison of representative graph transformers.
$\mathcal{N}_i$ denotes the sparse attention neighbourhood of node~$i$.
$\mathrm{PE}$ and $\mathrm{SE}$ stand for \emph{positional} and \emph{structural} encodings, respectively.
$^{\dagger}$The degree embedding in Graphormer is implemented as a \emph{learned} layer.
SAN: $p_i$ is a learned spectral positional encoding}
\label{tab:graph-transformer-comparison}
\end{table*}
\newline
\newline
One prominent example is \textbf{Exphormer} \cite{shirzad2023exphormer}, which constructs a sparse attention graph by combining three sources of connectivity: (i) the original graph edges \( E_{\text{orig}} \), (ii) additional random expander edges \( E_{\text{exp}} \), and (iii) virtual global nodes \( V_{\text{virt}} \). The resulting attention graph is defined as:
{
  \setlength{\abovedisplayskip}{2pt}%
  \setlength{\belowdisplayskip}{2pt}%
  \begin{equation}
\mathcal{G}_{\mathrm{attn}} = (V \cup V_{\mathrm{virt}},\, E_{\mathrm{orig}} \cup E_{\mathrm{exp}} \cup E_{\mathrm{virt}})
  \end{equation}
}
In this setup, each node attends only to a limited set of neighbors defined in the constructed attention graph. Expander edges form a fixed \( d \)-regular graph to ensure efficient long-range communication, while virtual nodes act as global hubs to help share information across the graph.

Let \( \mathcal{N}_i \subset V \cup V_{\text{virt}} \) denote the neighborhood of node \( i \) in \( \mathcal{G}_{\text{attn}} \). Then, attention is computed as (5):
{
  \setlength{\abovedisplayskip}{2pt}%
  \setlength{\belowdisplayskip}{2pt}%
  \begin{equation}
    \mathrm{Attention}_i = \mathrm{softmax}\left( \left\{ \tfrac{Q_i K_j^\top}{\sqrt{d_k}} \right\}_{j \in \mathcal{N}_i} \right) \cdot \left\{ V_j \right\}_{j \in \mathcal{N}_i}
  \tag*{}
  \stepcounter{equation}
  \end{equation}
}
Since each node attends to only a small, fixed number of others, the overall attention computation scales linearly. By combining local edges, sparse global links, and virtual tokens, Exphormer captures rich graph structure efficiently.

Another sparse approach is taken by \textbf{GraphGPS}~\cite{rampasek2022graphgps}, which adopts a hybrid architecture combining local message passing with global sparse attention.

Before processing, each node's features are augmented with positional $\mathrm{PE}_i$ (e.g., Laplacian eigenvectors) and structural $\mathrm{SE}_i$, (e.g., random walk features) encodings:
\vspace{-3pt}
\[
X_i \gets [X_i \, \| \, \mathrm{PE}_i \, \| \, \mathrm{SE}_i].
\]
The enriched features are then processed in alternating local and global stages. Each model block consists of:
\newline  
\textsc{Local step:} A GNN propagates information over the original graph edges \( E_{\text{orig}} \), updating node embeddings as  
\vspace{-3pt}
\[
X_i \gets \mathrm{GNNBlock}(X, E_{\mathrm{orig}})_i;
\]
\textsc{Global step:} Sparse attention is computed using a Transformer layer (e.g. Performer~\cite{choromanski2021rethinking}), where each node attends to a selected set \( \mathcal{N}_i \subset V \) (6):  
{
  \setlength{\abovedisplayskip}{2pt}%
  \setlength{\belowdisplayskip}{2pt}%
  \begin{equation}
\mathrm{Attention}_i = \mathrm{softmax}\left( \left\{ \tfrac{Q_i K_j^\top}{\sqrt{d_k}} \right\}_{j \in \mathcal{N}_i} \right) \cdot \left\{ V_j \right\}_{j \in \mathcal{N}_i}
  \tag*{}
  \stepcounter{equation}
  \end{equation}
}
The outputs of the two stages are \emph{combined additively}, integrating local and global information at each layer. 
\newline Since the choice of positional and structural encodings, GNN layers, and Transformer-based attention mechanisms is flexible, the architecture can be tailored to the specific task. This modular design decouples local and global processing, enabling GraphGPS to capture multi-scale dependencies efficiently while maintaining linear attention cost.
\begin{table*}[t]
  \centering
  \scriptsize
  \setlength{\tabcolsep}{4pt}
  \renewcommand{\arraystretch}{0.83}

  \begin{tabular*}{\textwidth}{@{\extracolsep{\fill}}lccccccc}
    \toprule
    \textbf{Dataset (Type)} & \textbf{Avg.\ Size} & \textbf{Task} &
    \textbf{Metric} & \textbf{Graphormer} & \textbf{SAN} &
    \textbf{GraphGPS} & \textbf{Exphormer} \\
    \midrule
    \textbf{ogbg-molhiv} (mol.) &
    $25$ n / 27 e &
    Binary cls. &
    ROC-AUC $\uparrow$ &
    \textbf{0.8051$\pm$0.0053} &
    0.7785$\pm$0.0247 &
    0.7880$\pm$0.0101 &
    --- \\[2pt]

    \textbf{ogbg-molpcba} (mol.) &
    $26$ n / 28 e &
    128-task cls. &
    AP $\uparrow$ &
    \textbf{0.3140$\pm$0.0032} &
    0.2765$\pm$0.0042 &
    0.2907$\pm$0.0028 &
    --- \\[2pt]

    \textbf{Peptides-func} (LRGB) &
    $151$ n / 307 e &
    Multi-label cls. &
    AP $\uparrow$ &
    --- &
    0.6439$\pm$0.0075 &
    \textbf{0.6535$\pm$0.0041} &
    0.6527$\pm$0.0043 \\[2pt]

    \textbf{ogbn-arxiv} (cit.) &
    169 k n / 1.16 M e &
    Node cls. &
    Acc. $\uparrow$ &
    OOM &
    OOM &
    OOM &
    \textbf{72.4$\pm$0.3} \\
    \bottomrule
  \end{tabular*}

  \captionsetup{skip=6pt}%
  \caption{GT benchmarks. ROC-AUC = area under the receiver operating characteristic curve, AP = average precision, Acc. = accuracy; $\uparrow$ = higher is better; OOM = out-of-memory, -- = not reported; $n$ = nodes, $e$ = edges.}
  \label{tab:benchmarks}
\end{table*}
\subsection{Empirical Comparison}
\label{sec:empirical}

Table~\ref{tab:benchmarks} shows results for Graphormer, SAN, GraphGPS, and Exphormer on benchmarks from the Open Graph Benchmark~\cite{hu2021ogb} and Long-Range Graph Benchmark (LRGB)~\cite{dwivedi2022lrgb}.

On small molecular graphs, dense-attention models achieve the strongest results. Graphormer~\cite{ying2021graphormer} leads on both \textit{ogbg-molhiv} and \textit{ogbg-molpcba}, with SAN~\cite{kreuzer2021rethinking} not far behind. However, their high memory requirements make them unsuitable for larger graphs: both fail on \textit{ogbn-arxiv} due to out-of-memory errors.

Sparse models, in contrast, scale more gracefully. GraphGPS~\cite{rampasek2022graphgps} performs well on molecular and long-range tasks, including the LRGB \textit{Peptides-func} dataset. Exphormer~\cite{shirzad2023exphormer} is the only model that runs on \textit{ogbn-arxiv}, reaching 72.4\% accuracy while remaining computationally efficient.
\section{Discussion \& Open Problems}
\label{sec:discussion}
\subsection{Trade-off Analysis}

The benchmark results show a clear trade-off between accuracy and scalability in graph transformer design. Dense attention, as used in Graphormer and SAN, performs best on small graphs like \textit{ogbg-molhiv} and \textit{molpcba}, where global context is important and computational cost is manageable. However, these models fail on large graphs like \textit{ogbn-arxiv} due to their quadratic memory requirements.

Sparse attention addresses this by reducing complexity to linear. Exphormer is the only model that runs on \textit{ogbn-arxiv}, showing that sparse designs are essential for large-scale graph processing. On medium-sized graphs such as \textit{Peptides-func}, sparse models like GraphGPS achieve competitive or even superior performance, showing that sparsity can be effective beyond just scaling.

Overall, the results show that the choice between dense and sparse attention should be guided by the graph’s size and structure. Flexible designs that adjust this balance can help models stay efficient without sacrificing performance, especially on graphs that fall between small and large scales.

\subsection{Limitations \& Theoretical Gaps}
While sparse attention mechanisms have shown practical promise, their expressive capacity remains far less well understood than that of dense Transformers, which are known to be universal approximators of sequence-to-sequence mappings~\cite{yun2020transformers}. Most sparse variants, such as Exphormer or GraphGPS, rely on fixed or heuristic sparsity patterns that may not generalize across different graph structures. In particular, the influence of design elements like expander edges, virtual tokens, or positional encodings on information flow and required depth is not yet fully characterized, leaving important open questions for future theoretical work~\cite{dwivedi2020benchmarking}.

\subsection{Future Directions}
Building on these open questions, future work should aim to establish stronger theoretical links between sparsity patterns and graph properties like conductance or spectral gap~\cite{dwivedi2021generalization}. Dynamic sparsification, as in NodeFormer~\cite{wu2022nodeformer}, is another promising approach for learning which edges to keep during training. Finally, large-scale tests on web-scale benchmarks like OGB and LRGB~\cite{hu2021ogb,dwivedi2022lrgb} are essential to assess real-world scalability and will help bridge the gap between theory and practical usage.

\section{Conclusion}
We presented a comparative analysis of dense and sparse attention mechanisms in graph Transformers. Dense attention models-such as Graphormer and SAN-excel at capturing global dependencies but face scalability limits due to their quadratic time and memory complexity. In contrast, sparse architectures-like GraphGPS and Exphormer—achieve linear complexity while maintaining strong performance, especially on large graphs.

Answering our central question: dense attention remains preferable for small to medium graphs where full context justifies the cost, while sparse attention is essential for large-scale graphs where efficiency and reach must be balanced. Future graph Transformers must combine theoretical insights, adaptive sparsification, and robust benchmarks to navigate this trade-off between expressivity and scalable complexity.
\newpage
\bibliographystyle{icml2024}
\bibliography{references}
\end{document}